\documentclass[letterpaper]{article} % DO NOT CHANGE THIS
\usepackage{aaai2026}  % DO NOT CHANGE THIS
\usepackage{times}  % DO NOT CHANGE THIS
\usepackage{helvet}  % DO NOT CHANGE THIS
\usepackage{courier}  % DO NOT CHANGE THIS
\usepackage[hyphens]{url}  % DO NOT CHANGE THIS
\usepackage{graphicx} % DO NOT CHANGE THIS
\usepackage{natbib}  % DO NOT CHANGE THIS AND DO NOT ADD ANY OPTIONS TO IT
\usepackage{caption} % DO NOT CHANGE THIS AND DO NOT ADD ANY OPTIONS TO IT
\usepackage{algorithm}
\usepackage{algorithmic}

\usepackage{newfloat}
\usepackage{listings}
\DeclareCaptionStyle{ruled}{labelfont=normalfont,labelsep=colon,strut=off} % DO NOT CHANGE THIS
\floatstyle{ruled}
\newfloat{listing}{tb}{lst}{}
\floatname{listing}{Listing}
\usepackage{amsmath}
\DeclareMathOperator*{\argmin}{arg\,min}
\usepackage{amsfonts}
\usepackage{tikz}
\usepackage{pgfplots}
\usetikzlibrary{patterns}
\usepackage{amssymb}
\usepackage{subcaption}
\usepackage{multirow}

\nocopyright
\title{Leveraging Intermediate Representations of Time Series Foundation Models for Anomaly Detection}
\author{
    Chan Sik Han,
    Keon Myung Lee\thanks{Keon Myung Lee is the corresponding author.}
}
\affiliations{
    Chungbuk National University\\
    \{chatterboy,kmlee\}@chungbuk.ac.kr
}

\usepackage{bibentry}
\begin{document}

\maketitle

\begin{abstract}
Detecting anomalies in time series data is essential for the reliable operation of many real-world systems. Recently, time series foundation models (TSFMs) have emerged as a powerful tool for anomaly detection. However, existing methods typically rely on the final layer's representations of TSFMs, computing the anomaly score as a reconstruction or forecasting error via a task‑specific head. Instead, we propose TimeRep, a novel anomaly detection approach that leverages the intermediate layer's representations of TSFMs, computing the anomaly score as the distance between these representations.
Given a pre-trained TSFM, TimeRep selects the intermediate layer and patch‑token position that yield the most informative representation.
TimeRep forms a reference collection of intermediate representations from the training data and applies a core-set strategy to reduce its size while maintaining distributional coverage.
During inference, TimeRep computes the anomaly score for incoming data by measuring the distance between its intermediate representations and those of the collection.
To address concept drift, TimeRep integrates an adaptation mechanism that, at inference time, augments the collection exclusively with non-redundant intermediate representations from incoming data.
We conducted extensive experiments on the UCR Anomaly Archive, which contains 250 univariate time series. TimeRep consistently outperforms a broad spectrum of state-of-the-art baselines, including non-DL, DL, and foundation model-based methods.
\end{abstract}


\section{Introduction}

Unexpected behaviors in real-world systems frequently appear in subtle forms, such as unusual vibrations in industrial machinery, slight latency in cloud services, or minor irregularities in physiological signals captured by wearable devices. Although these anomalies may initially seem negligible, they often serve as early indicators of critical failures, system outages, or medical emergencies. Hence, timely and accurate anomaly detection is essential to ensure operational safety, improve system reliability, and reduce costs in various domains, including manufacturing, healthcare, finance, and IT infrastructure.

Most modern systems generate time series data, comprising observations from sensors, system logs, and transactions. This time series data serve as a primary means of monitoring system behavior and identifying abnormal patterns. However, despite the abundance of observational data, reliable anomaly detection remains challenging due to the rarity, diversity, and contextual nature of anomalies. For instance, a sudden spike may indicate a malfunction in one setting but reflect normal variation in another. These challenges are further exacerbated by the non-stationary nature of real-world environments, where the definition of normality evolves over time due to seasonal trends, system modifications, and shifting user behavior, a phenomenon widely recognized as concept drift.

Prior work has explored a variety of approaches to address these challenges, including rule-based heuristics, statistical models, and traditional machine learning methods. However, these conventional techniques often lack the flexibility to capture the complexity and non-stationarity of real-world time series data. Deep learning models have shown greater promise by learning non-linear and contextual patterns, but they remain sensitive to distribution shifts and typically require frequent retraining. Therefore, both traditional and modern approaches often fail to generalize across domains or adapt to evolving environments, leading to missed anomalies or false alarms, particularly under distribution shifts or in previously unseen scenarios.

Recently, time series foundation models (TSFMs) have emerged as a promising direction for generalizable anomaly detection. These models learn general-purpose temporal representations from large and diverse time series corpora, allowing for their ready adaptation to a wide range of downstream tasks \cite{das2024decoder,ansari2024chronos,woo2024unified,shi2025timemoe,liu2025sundial}.
However, existing TSFM-based methods typically utilize final layer representations via a task-specific head, such as a reconstruction or forecasting head, for the anomaly detection task \cite{zhou2023one,liu2024timer,zhang2025timesbert}.
This design suffers from three major limitations.
First, final layer representations often overfit to the pretraining objective, limiting their effectiveness for the anomaly detection task.
Second, task-specific heads often fail in practice because reconstruction heads tend to reproduce anomalies too faithfully and forecasting heads struggle when future patterns are unstable or unpredictable.
Third, most TSFM-based methods lack an adaptation mechanism for handling concept drift at inference time. Although a recent study \cite{kim2024model} proposes test-time adaptation for unsupervised time series anomaly detection, it involves updating model parameters during inference, which limits its practicality in real-world deployment settings.

Here, we propose TimeRep, an unsupervised anomaly detection method to address the aforementioned challenges. TimeRep directly utilizes intermediate layer representations from a TSFM for anomaly detection, instead of using final layer representations via a task-specific head. During training, TimeRep extracts patch-level intermediate layer representations and stores them in a reference collection, which serves as a memory bank. To improve efficiency, TimeRep compresses the memory bank using a core-set strategy that maintains representative coverage of stored representations. At inference time, TimeRep detects anomalies by computing the distance between the intermediate layer representations of incoming time series data and those stored in the memory bank. The anomaly score is defined as the distance to the nearest item in the memory, where a greater distance indicates a higher likelihood of anomaly. To handle concept drift, TimeRep updates the memory bank at inference time by adding non-redundant intermediate layer representations of incoming data. This enables TimeRep to adapt to evolving patterns of time series data without updating the model parameters.

We summarize our contributions as follows:
\begin{itemize}
    \item We propose TimeRep, a simple yet effective anomaly detection method that directly leverages intermediate layer representations from a time series foundation model, with no need of fine-tuning on individual datasets.
    \item TimeRep integrates a test-time adaptive memory bank that dynamically incorporates non-redundant intermediate representations of incoming data at inference time, allowing the model to maintain performance under concept drift without modifying its parameters.
    \item We demonstrate the effectiveness of TimeRep through extensive experiments on the UCR Anomaly Archive, including 250 diverse univariate time series, comparing it against state-of-the-art methods spanning non-DL, DL, and foundation model-based approaches.
\end{itemize}

\section{Related Work}

\subsection{Time Series Anomaly Detection}

Time series anomaly detection (TSAD) has evolved from rule-based heuristics to deep learning and foundation model-based approaches. For an in-depth survey of this progression, see a recent work \cite{qiu2025tab}.
Early TSAD methods include rule-based systems and statistical models like ARIMA, STL, and CUSUM, valued for their interpretability. Distance-based approaches have also gained attention, notably the Matrix Profile \cite{yeh2016matrix}, which detects discords via minimum similarity to other subsequences. Recently, the DAMP \cite{lu2022matrix} extended discord-based anomaly detection by introducing a backward iterative doubling search and forward pruning scheme, achieving state-of-the-art results without deep learning.
With the rise of deep learning, attention-based models have become increasingly dominant. The Anomaly Transformer \cite{xu2022anomaly} introduces association discrepancy to localize abnormal time steps, while the NPSR \cite{lai2023nominality} proposes a nominality-aware reconstruction objective. The SensitiveHUE \cite{feng2024sensitivehue} incorporates predictive uncertainty to improve robustness, and the CATCH \cite{wu2025catch} detects anomalies by patching time series in the frequency domain and using masked attention to model channel-wise correlations.
More recently, researchers have increasingly adopted large-scale foundation models (FMs), including both pre-trained language models and time series models. Many approaches utilize final layer representations coupled with task-specific heads, such as a reconstruction or forecasting head, which require fine-tuning on downstream tasks to enhance performance. Notable examples include TimeGPT-1 \cite{garza2023timegpt}, GPT4TS \cite{zhou2023one}, Timer \cite{liu2024timer}, MOMENT \cite{goswami2024moment}, TimesBERT \cite{zhang2025timesbert}, and DADA \cite{shentu2025towards}. In contrast, AnomalyLLM \cite{liu2024large} leverages intermediate layer representations from a GPT-2 model and employs a teacher–student framework that involves dataset-specific fine-tuning for both networks. Due to representation inconsistency between deep teacher layers and the student network, such models often restrict representation extraction to shallower layers.
To overcome these limitations, the proposed method performs anomaly detection by directly leveraging intermediate layer representations from a time series foundation model, without relying on any task-specific heads or requiring dataset-level fine-tuning. We observe that these intermediate representations encode highly effective temporal structures for the anomaly detection task.

\begin{figure*}[!ht]
\centering
\includegraphics[width=0.85\textwidth]{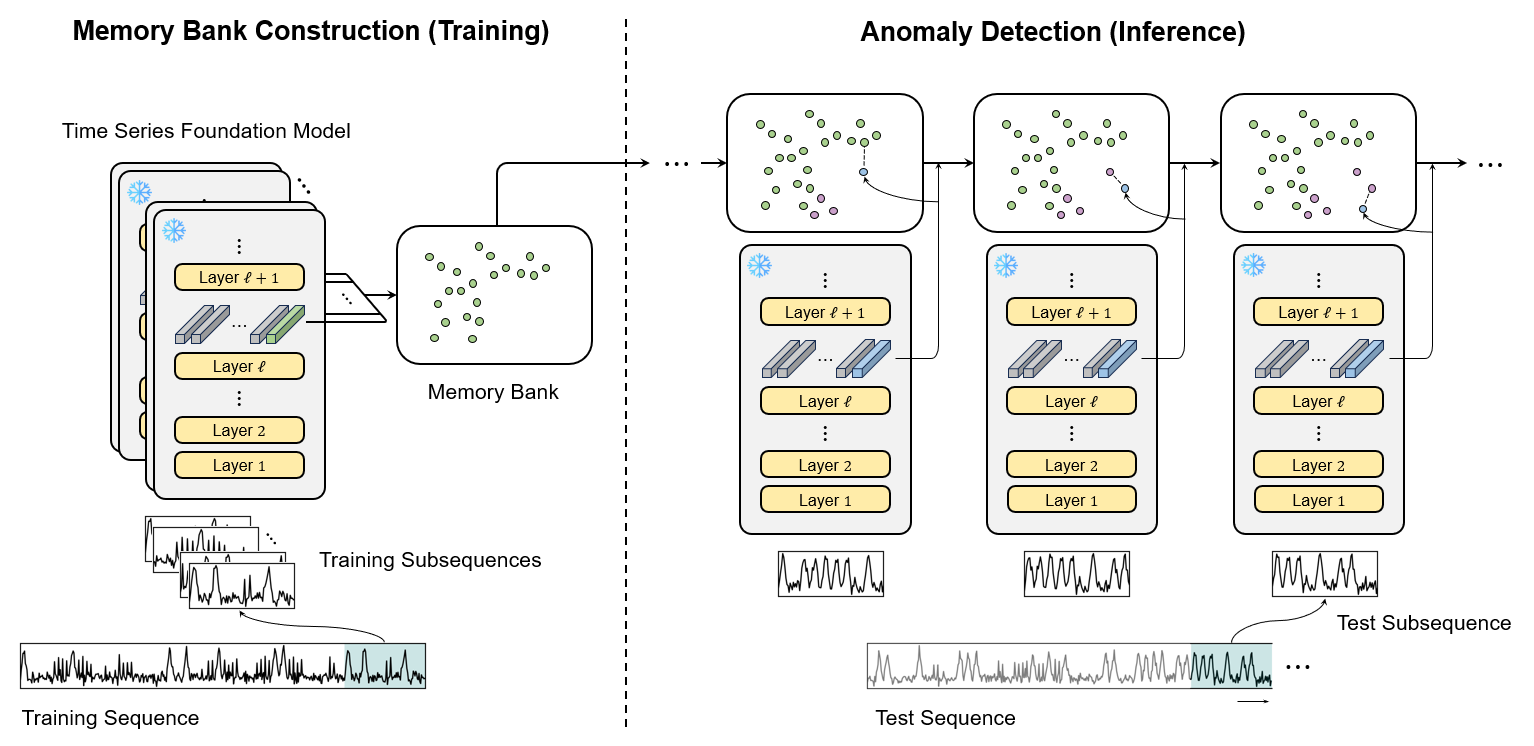} % Reduce the figure size so that it is slightly narrower than the column.
\caption{Overview of the proposed anomaly detection method. During training, intermediate patch representations extracted from a time series foundation model and then are stored in a memory bank. At inference, the intermediate patch representation of each incoming data is compared against the bank to compute its anomaly score, and non-redundant representations of new observations is dynamically added to the bank to adapt to concept drift.}
\label{fig1}
\end{figure*}

\subsection{Internal Representations of Foundation Models}

Internal representations are key to understanding how deep learning models encode information. Early work \cite{alain2016understanding} showed that different layers capture varying abstraction levels. With foundation models, interest has grown in internal representations for generalization, robustness, and interpretability.
In NLP, representational similarity metrics such as CKA and RSA have been used to analyze the structural properties of internal representations across architectures \cite{brown2023understanding} and techniques like activation patching have linked internal states to safety-related outputs \cite{yamaguchi2025adversarial}. Internal representation quality has been tied to downstream performance \cite{skean2025layer}, and aligned multilingual embeddings have improved cross-lingual transfer \cite{liu2025middle}.
In computer vision and multimodal learning, internal features have been used to analyze semantic alignment \cite{lee2025multimodal} and their relationship to classification accuracy \cite{skean2025layer}.
For time series foundation models (TSFMs), recent studies have analyzed representational similarity, introduced decoding and interventions, and assessed latent space structure \cite{wilinski2025exploring, santamaria2025decoding}.
While prior work has focused on analyzing internal representations of TSFMs, we directly leverage them for the anomaly detection task in time series data.

\section{Proposed Method}

\subsection{Problem Formulation}

This study addresses unsupervised anomaly detection in univariate time series data. Let the time series be denoted as $\mathcal{X}=\{x_{1},x_{2},\cdots,x_{T}\}$, where $x_{t}\in\mathbb{R}$ is a real-valued observation at time step $t$ and $T$ is the sequence length. The dataset consists of a training set and a test set. The training set contains normal data without labels, while the test set includes both time series values and binary anomaly labels $\mathcal{Y}=\{y_{1},y_{2},\cdots,y_{T}\}$, where each $y_{t}\in\{0,1\}$ indicates whether time step $t$ is anomalous. To prepare the data for modeling, a sliding window of length $L$ is applied to generate overlapping subsequences. Each subsequence $\mathbf{s}_{t}=\{x_{t-L+1},\cdots,x_{t-1},x_{t}\}\in\mathbb{R}^{L}$ is extracted by shifting the window forward by one time step. These subsequences are fed into an anomaly detection model $f(\cdot)$, which produces a score $a_{t}$ for each time step. During inference, anomaly scores $\{s_{1},s_{2},\cdots,s_{T}\}$ are computed for the test sequence. Each time step is classified as anomalous when its score $s_{t}$ exceeds a predefined threshold $\theta$. The predicted labels $\{\hat{y}_{1},\hat{y}_{2},\cdots,\hat{y}_{T}\}$ are then compared with the ground truth labels $\mathcal{Y}$ using standard evaluation metrics.

% This study addresses unsupervised anomaly detection in univariate time series data. Let the univariate time series data be denoted as $\mathcal{X}=\{x_{1},x_{2},\cdots,x_{T}\}$, where each $x_{t}\in\mathbb{R}$ represents a real-valued observation at time step $t$ and $T$ is the total length of the sequence. The dataset is composed of a training set and a test set. The training set contains normal data and does not include any labels. In contrast, the test set includes both the time series values and the corresponding ground-truth binary anomaly labels $\mathcal{Y}=\{y_{1},y_{2},\cdots,y_{T}\}$, where each $y_{t}\in\{0,1\}$ indicates whether the time step $t$ is anomalous or normal. The sliding window technique is applied to prepare the time series for modeling by generating overlapping fixed-length subsequences. For a given window length $L$, each subsequence $\mathbf{s}_{t}=\{x_{t-L+1},\cdots,x_{t-1},x_{t}\}\in\mathbb{R}^{L}$ is extracted by shifting the window forward by one time step. Each subsequence $\mathbf{s}_{t}$ is then fed into an anomaly detection model $f(\cdot)$, which produces an output used to compute an anomaly score $a_{t}$. During inference on the test set, this process is applied to the entire sequence, resulting in a sequence of anomaly scores $\{a_{1},a_{2},\cdots,a_{T}\}$. Each time step is subsequently classified as anomalous or normal based on whether the corresponding score exceeds a predefined threshold $\theta$. Finally, the predicted binary labels are compared with the ground-truth labels to evaluate detection performance using metrics.

\subsection{Pre-Trained Time Series Foundation Model}

In this study, we leverage the representational power of a time series foundation model (TSFM) trained on large-scale time series data by utilizing intermediate layer representations for anomaly detection. Specifically, we adopt the MOMENT-Large model \cite{goswami2024moment}, an encoder-only TSFM pre-trained on the Time Series Pile dataset, which comprises over 1.13 billion observations from diverse domains. The model consists of 24 layers, 16 attention heads, and a model dimension of 1024, totaling approximately 385 million parameters. It processes input subsequences in patch units to capture local temporal dependencies. In our approach, we utilize intermediate layer representations extracted by the model for anomaly detection.

% The objective of this study is to effectively leverage the representational power of a time series foundation model pre-trained on large-scale time series data. In this work, we utilize MOMENT-Large \cite{goswami2024moment}, a recently proposed encoder-only time series foundation model. MOMENT-Large is pre-trained on the Time Series Pile dataset, which consists of 1.13B observations. In addition, MOMENT-Large has approximately 385 million parameters and consists of 24 layers. The model dimension is 1024, and it utilizes 16 attention heads. Input time series segments are processed in patch units by the model.In this study, for anomaly detection, we utilize the feature representation at the $n$-th position of the $l$th layer, given the input time series segment at time $t$, $\mathbf{r}_{t,n}^{l}\in\mathbb{R}^{d_{model}}$.

\subsection{Memory-based Anomaly Detection}

Figure \ref{fig1} presents the proposed anomaly detection method that leverages intermediate layer representations extracted from a time series foundation model. During training, representations of training subsequences are stored in the memory bank. At inference time, representations of incoming subsequences are compared against memory items in the memory bank to compute anomaly scores based on their distance from known patterns.

\paragraph{Patch Representation}

Given an input subsequence $\mathbf{s}_{t}$, the TSFM divides it into $N$ non-overlapping patches and encodes each into a representation at every layer. From the $l$-th layer, we obtain a set of patch representations $\{\mathbf{r}_{t,1}^{(l)},\mathbf{r}_{t,2}^{(l)},\cdots,\mathbf{r}_{t,N}^{(l)}\}$, where $\mathbf{r}_{t,n}^{(l)}\in\mathbb{R}^{d_{\text{model}}}$ denotes the $n$-th patch representation. Each serves as a compact semantic summary of its segment, capturing both fine-grained temporal features and broader contextual patterns. We associate each subsequence $\mathbf{s}_{t}$ with a specific reference time step within it, so that the anomaly score computed from its representation reflects that particular time step. We consider two settings for this reference position. In the first setting, it corresponds to the final time step of the subsequence. In the second, it corresponds to the temporal center. In both settings, the representation is extracted from a specific layer of the TSFM. The choice of layer and reference time step was guided by empirical analysis, as discussed in Sections 4.3 and 4.5.

% Given an input subsequence $\mathbf{s}_{t}$ at time step $t$, the TSFM divides it into $N$ non-overlapping patches and encodes each patch into a representation at every layer. From the $l$-th layer, we obtain a set of patch representations $\{\mathbf{r}_{t,1}^{(l)}, \mathbf{r}_{t,2}^{(l)},\cdots,\mathbf{r}_{t,N}^{(l)}\}$, where $\mathbf{r}_{t,n}^{(l)}\in\mathbb{R}^{d_{\text{model}}}$ denotes the representation of the $n$-th patch. Each patch representation serves as a compact semantic summary of its corresponding segment, capturing both fine-grained temporal characteristics and broader contextual patterns. Here, each input subsequence $\mathbf{s}_t$ is associated with a specific reference time step within the subsequence. We consider two configurations for this reference position. One corresponds to the final time step in the subsequence, and the other corresponds to its temporal center. In both settings, we extract patch representations from the 16th layer of the TSFM. The impact of this choice is analyzed in our ablation study (Section 4.3 and 4.5).

\paragraph{Memory Bank Construction}

To detect anomalies based on semantic similarity, we construct a memory bank that stores intermediate representations extracted from the training data. For each subsequence $\mathbf{s}_{t}$, we obtain its corresponding representation $\mathbf{r}_{t}$ using the TSFM, as described in Section 3.3.1. These representations, reflecting normal patterns, form a memory bank $\mathcal{M}=\{\mathbf{m}_{1},\mathbf{m}_{2},\cdots,\mathbf{m}_{K}\}$. This memory serves as a reference set against which incoming test representations are compared to assess deviation from normality.

\paragraph{Enhancing Memory Bank Efficiency}

Retaining all intermediate representations from the training data can quickly lead to a large and inefficient memory bank. This not only increases storage requirements and inference time, but also introduces redundancy that may degrade anomaly detection accuracy. To overcome these challenges, we employ a coreset selection method that compresses the memory while preserving the diversity of normal patterns.

We adopt the k-Center selection method for memory compression by selecting a subset that best covers the distribution of training representations. The objective is to minimize the maximum distance between any point in the original memory bank and its closest selected representative, thereby ensuring uniform coverage across the training distribution. This can be formally stated as the following optimization problem:
\begin{equation}
\mathcal{C}^{*}=\argmin_{\mathcal{C}\subset\mathcal{M}}\max_{\mathbf{m}\in\mathcal{M}}\min_{\mathbf{c}\in\mathcal{C}}\|\mathbf{m}-\mathbf{c}\|_{2}
\end{equation}
where $\mathcal{M}$ denotes the full set of training-time representations, and $\mathcal{C}$ is the coreset to be selected. This optimization guarantees that the selected subset provides broad and representative coverage of the training distribution. However, solving this problem exactly is known to be NP-hard, which makes it computationally infeasible for large-scale time series datasets. To address this, we apply a greedy approximation of the k-Center algorithm \cite{sener2018active}. The algorithm starts by randomly selecting an initial representation and then iteratively adds the point that is farthest from the current set. This process continues until the desired memory size is reached, resulting in a compact yet diverse coreset that effectively captures the structure of the training data.

\begin{table*}[!ht]
\centering
\renewcommand{\arraystretch}{1.2}
%\resizebox{.95\columnwidth}{!}{
\begin{tabular}{c|c|c|c|c}
    \hline
    \textbf{Approach} & \textbf{Model Type} & \textbf{Evaluation Setting} & \textbf{Method} & \textbf{Top-1 Accuracy} \\
    \hline
    \multirow{4}{*}{non-DL} & \multirow{4}{*}{Specialized} & \multirow{4}{*}{from scratch} & IForest & 37.6 \\
    & & & SCRIMP & 41.6 \\
    & & & NORMA & 47.4 \\
    & & & DAMP & 63.2 \\
    \hline
    \multirow{11}{*}{DL} & \multirow{5}{*}{Specialized} & \multirow{5}{*}{from scratch} & LSTM-VAE & 19.8 \\
    & & & AE & 23.6 \\
    & & & USAD & 27.6 \\
    & & & COCA & 27.6 \\
    & & & TimeVQVAE-AD & 66.8 \\
    \cline{2-5}
    & \multirow{6}{*}{Foundation} & \multirow{2}{*}{fine-tuning} & MOMENT & 32.8 \\
    & & & Timer & 36.0 \\
    \cline{3-5}
    & & \multirow{4}{*}{w/o fine-tuning} & MOMENT & 23.2 \\
    & & & Timer & 20.0 \\
    \cline{4-5}
    & & & Ours (last) & \textbf{72.8} \\
    & & & Ours (center) & \textbf{77.6} \\
    \hline
\end{tabular}
\caption{Performance comparison on the UCR Anomaly Archive. The methods are grouped by approach such as non-DL and DL, by model type such as specialized and foundation, and by evaluation setting including from scratch, fine-tuning, and without fine-tuning. Our methods demonstrate the superior performance against all other methods.}
\label{tab1}
\end{table*}

\paragraph{Distance-based Anomaly Scoring}

The anomaly score is computed as the distance between the intermediate representation of an input subsequence and its closest match in the memory bank. Representations that deviate significantly from all stored patterns are assigned higher scores, reflecting a greater likelihood of anomaly. Formally, let $\mathbf{r}_{t}$ denote the intermediate representation at time step $t$ and let $\mathcal{M}$ be the set of stored representations in the memory bank. The anomaly score $s_{t}$ is computed as:
\begin{equation}
s_{t}=d(\mathbf{r}_{t},\mathbf{m}_{t}^{*})
\end{equation}
where the nearest neighbor $\mathbf{m}_{t}^{*}$ is defined as:
\begin{equation}
\mathbf{m}_{t}^{*}=\argmin_{\mathbf{m}\in\mathcal{M}}\|\mathbf{r}_{t}-\mathbf{m}\|_{2}
\end{equation}

We consider three types of distance functions for computing the anomaly score. Each offers different assumptions and sensitivities to data distribution. The Euclidean distance computes the straight-line distance between the input representation $\mathbf{r}_{t}$ and its nearest memory item $\mathbf{m}_{t}^{*}$. It assumes a uniform feature space without correlation across dimensions:
\begin{equation}
  d_{E}(\mathbf{r}_{t},\mathbf{m}_{t}^{*})=\| \mathbf{r}_{t}-\mathbf{m}_{t}^{*}\|_{2}
\end{equation}

The Mahalanobis distance incorporates the covariance structure $\Sigma$ of the memory bank, allowing the distance to adapt to the anisotropy in the feature space. It is defined as:
\begin{equation}
  d_{M}(\mathbf{r}_{t},\mathbf{m}_{t}^{*};\Sigma)=\sqrt{(\mathbf{r}_{t}-\mathbf{m}_{t}^{*})^{T}\Sigma^{-1}(\mathbf{r}_{t}-\mathbf{m}_{t}^{*})}
\end{equation}

Following \cite{roth2022towards}, we define a density-aware distance that adjusts the anomaly score by incorporating local neighborhood density. This formulation penalizes points that lie in low-density regions of the memory bank, treating them as more likely to be anomalous even when they are close to a neighbor. The score is defined as:
\begin{multline}
d_{N}(\mathbf{r}_{t}, \mathbf{m}_{t}^{*}) = \left(1 - 
\frac{e^{\| \mathbf{r}_{t} - \mathbf{m}_{t}^{*} \|_{2}}}
{\sum_{\mathbf{n} \in \mathcal{N}_{b}(\mathbf{m}_{t}^{*})} 
e^{\| \mathbf{r}_{t} - \mathbf{n} \|_{2}}} \right) \\
\times \| \mathbf{r}_{t} - \mathbf{m}_{t}^{*} \|_{2}
\end{multline}
where $\mathcal{N}_{b}(\mathbf{m}_{t}^{*})$ denotes the set of $b$ nearest neighbors of $\mathbf{m}_{t}^{*}$ in the memory bank. To improve numerical stability, we apply max-subtraction before the exponential computation by subtracting the largest distance in the denominator. The resulting values are then scaled by the model dimensionality, $\sqrt{d_{\text{model}}}$.

\subsection{Test-Time Adaptive Memory Bank}

In time series anomaly detection, the underlying patterns of the data often shift after training, a phenomenon known as concept drift. Without proper adaptation, such changes can lead to a significant drop in anomaly detection performance during inference. To address this, we freeze the model and instead adapt the memory bank by selectively adding intermediate representations from incoming test subsequences. To implement this selective update, we introduce a novelty-based filtering criterion to avoid adding redundant representations to the memory bank. Specifically, we compute the distance between the intermediate representation of incoming test subsequence to its nearest neighbor in the current memory bank. If the distance exceeds a predefined novelty threshold, the representation is considered sufficiently different and is inserted into the memory bank. Note that this novelty threshold is distinct from the anomaly detection threshold. We determine the novelty threshold by computing the nearest neighbor distance of each training representation for the memory bank and constructing the empirical cumulative distribution of these distances. The 80th percentile of this distribution is used as the threshold, ensuring that only sufficiently non-redundant representations are incorporated. This value remains fixed throughout all our experiments.

\section{Experiments}

\subsection{Experiment Settings}

\paragraph{Dataset}

Several benchmark datasets for univariate time series anomaly detection, such as Yahoo \cite{laptev2015yahoo}, NAB \cite{ahmad2017unsupervised}, NASA \cite{hundman2018detecting}, and IOPS \cite{ren2019time}, have been widely used in previous studies. However, recent work has pointed out that these benchmarks exhibit multiple flaws that negatively affect the reliability of model evaluation. In particular, prior work \cite{wu2023current} demonstrates that these issues can lead to misleading conclusions about model performance. To address these concerns, we adopt the UCR Anomaly Archive, a benchmark suite introduced in their study. It includes 250 univariate time series collected from both real-world and synthetic sources. The dataset covers a broad spectrum of domains such as cardiology, industrial systems, medicine, zoology, meteorology, and human activity.

\paragraph{Baselines}

To evaluate the effectiveness of the proposed model, we compare it against widely used and state-of-the-art baselines, including both traditional and DL-based approaches. The traditional methods include IForest \cite{liu2008isolation}, SCRIMP \cite{zhu2018matrix}, NORMA \cite{boniol2021unsupervised}, and DAMP \cite{lu2022matrix}. Among DL-based methods, we differentiate between methods that are tailored to individual datasets and those that leverage TSFMs. The specialized models include LSTM-VAE \cite{park2018multimodal}, AE \cite{audibert2021univariate}, USAD \cite{audibert2021univariate}, COCA \cite{wang2023deep}, and TimeVQVAE-AD \cite{lee2024explainable}. In contrast, TSFM-based methods include Timer \cite{liu2024timer}, MOMENT \cite{goswami2024moment}, and DADA \cite{shentu2025towards}.

\paragraph{Evaluation Metric}

Various metrics have been proposed for time series anomaly detection. To ensure fair and consistent evaluation, we adopt an evaluation metric that is both threshold-agnostic and hyperparameter-free. Although commonly used metrics such as AUC-ROC and AUC-PR meet these criteria, they often produce misleading results in settings with severe class imbalance, which is a common characteristic of anomaly detection, and when results are aggregated across diverse datasets.
To address these limitations, we adopt Top-1 accuracy \cite{wu2023current} as our primary evaluation metric, which is computed as follows. Suppose the evaluation includes $N$ distinct test datasets. For the $i$-th test dataset, we identify the timestamp with the highest anomaly score, denoted as $t_{i}^{*}$. A prediction is considered correct if it falls within a predefined tolerance window surrounding the ground truth anomaly region:
\begin{equation}
    \delta_i =
    \begin{cases}
    1, & \text{if } t_i^* \in [\,a_i - \Delta,\; b_i + \Delta\,] \\
    0, & \text{otherwise}
    \end{cases}
\end{equation}
Here, $a_i$ and $b_i$ denote the start and end timestamps of the ground truth anomaly in the $i$-th test dataset, and $\Delta$ is a fixed tolerance margin. Following prior work, we set $\Delta=100$. Top-1 accuracy is then computed as the proportion of test datasets in which the model successfully hits the anomaly region.

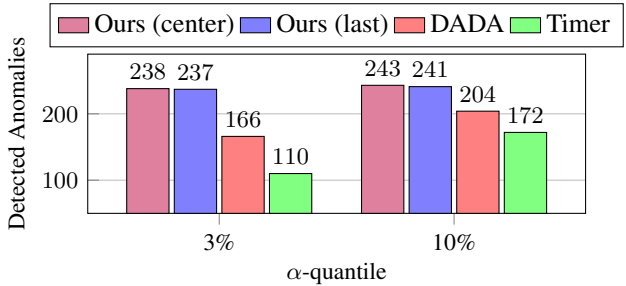
\begin{figure}
% \centering
\begin{tikzpicture}
\begin{axis}[
    ybar,
    bar width=16pt,  % 18pt
    width=8.2cm,
    height=3.7cm,
    ymin=50, ymax=290,
    ytick={100, 200},
    ylabel={Detected Anomalies},
    xlabel={$\alpha$-quantile},
    symbolic x coords={3\%, 10\%},
    xtick=data,
    xtick pos=bottom,
    ytick pos=left,
    legend style={at={(0.5,1.32)}, anchor=north, legend columns=4},
    legend image code/.code={
        \draw[draw=black]
        (0cm,-0.1cm) rectangle (0.4cm,0.1cm);
    },
    ylabel style={yshift=-0.8em},
    tick label style={font=\small},
    label style={font=\small},
    nodes near coords,
    nodes near coords align={vertical},
    enlarge x limits=0.56,
    grid=major,
    major x grid style={draw=none},
    nodes near coords,
    nodes near coords style={font=\small}
]
\addplot[
    draw=black, fill=purple!50,
] coordinates {(3\%, 238) (10\%, 243)}; \addlegendentry{Ours (center)}
\addplot[
    draw=black, fill=blue!50,
] coordinates {(3\%, 237) (10\%, 241)}; \addlegendentry{Ours (last)}
\addplot[
    draw=black, fill=red!50,
] coordinates {(3\%, 166) (10\%, 204)}; \addlegendentry{DADA}
\addplot[
    draw=black, fill=green!50,
] coordinates {(3\%, 110) (10\%, 172)}; \addlegendentry{Timer}
\end{axis}
\end{tikzpicture}
\caption{$\alpha$-quantile evaluation results for $3\%$ and $10\%$. Each bar shows the number of datasets with at least one anomaly detected within the top-$\alpha$ scored timestamps.}
\label{fig2}
\end{figure}

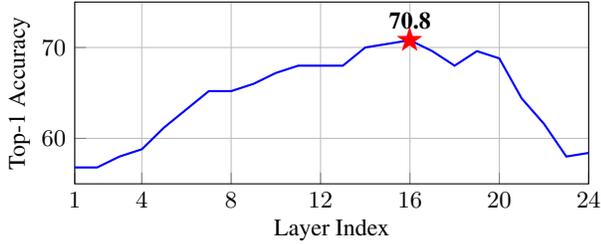
\begin{figure}[!]
\centering
\begin{tikzpicture}
\begin{axis}[
    height=4cm,
    width=\columnwidth,
    xlabel={Layer Index},
    ylabel={Top-1 Accuracy},
    ylabel style={yshift=-1.4em},
    xlabel style={yshift=0.5em},
    xtick={1,4,8,12,16,20,24},
    ymin=55, ymax=75,
    grid=major,
    enlarge x limits=0,
    ytick distance=10,
    outer sep=0pt,
    tick label style={font=\small},
    label style={font=\small},
    xtick pos=bottom,
    ytick pos=left
]
\addplot[
    thick,
    color=blue,
] coordinates {
    (1,56.8) (2,56.8) (3,58.0) (4,58.8) (5,61.2) (6,63.2)
    (7,65.2) (8,65.2) (9,66.0) (10,67.2) (11,68.0) (12,68.0)
    (13,68.0) (14,70.0) (15,70.4) (16,70.8) (17,69.6) (18,68.0)
    (19,69.6) (20,68.8) (21,64.4) (22,61.6) (23,58.0) (24,58.4)
};
\node at (axis cs:16,70.8) {\textcolor{red}{\large$\bigstar$}};
\node at (axis cs:16,73.0) {\textbf{\small 70.8}};
\end{axis}
\end{tikzpicture}
\caption{Layer-wise Top-1 accuracy for the evaluated model. Each data point represents the Top-1 accuracy obtained when the output of the corresponding layer is used for anomaly detection.}
\label{fig3}
\end{figure}

\subsection{Performance Comparison Evaluation}

% The evaluation follows the most common scenario in time series anomaly detection, where a new observation arrives at each time step and anomaly detection is performed continuously. To simulate this setting, we set the window stride to 1. Table \ref{tab1} presents the Top-1 accuracy results for all baseline methods on the UCR Anomaly Archive. The proposed method outperforms all baselines in both configurations.
% The last-aligned variant achieves 72.8\%, while the center-aligned variant achieves the highest score of 77.6\%, demonstrating the effectiveness of the proposed method across diverse datasets. Notably, this is achieved without any fine-tuning or task-specific head, and surpasses MOMENT—the strongest baseline under the same setting—which achieves 23.2\%.
% Figure \ref{fig2} shows the results of the $\alpha$-quantile evaluation for 3\% and 10\%. This metric assesses whether each method successfully ranks at least one true anomaly within the top $\alpha$ fraction of timestamps, based on anomaly scores. The proposed method outperforms baselines under both 3\% and 10\%. The center- and last-aligned variants detect over 240 datasets, while DADA and Timer detect significantly fewer. These results indicate superior anomaly ranking in a threshold-free setting.

The evaluation follows the most common scenario in time series anomaly detection, where a new observation arrives at each time step and anomaly detection is performed continuously. To simulate this setting, we set the window stride to 1. Table \ref{tab1} presents the Top-1 accuracy results for all baseline methods on the UCR Anomaly Archive. The proposed method outperforms all baselines in both configurations.
The last-aligned variant achieves 72.8\%, while the center-aligned variant achieves the highest score of 77.6\%, demonstrating the effectiveness of the proposed method across diverse datasets. Figure \ref{fig2} shows the results of the $\alpha$-quantile evaluation for 3\% and 10\%. This metric assesses whether each method successfully ranks at least one true anomaly within the top $\alpha$ fraction of timestamps, based on anomaly scores. The proposed method outperforms baselines under both 3\% and 10\%.  The center- and last-aligned variants detect over 240 datasets, while DADA and Timer detect significantly fewer. These results indicate superior anomaly ranking in a threshold-free setting.

\subsection{Layer-wise Performance Evaluation}

Figure~\ref{fig3} presents the Top-1 accuracy when using representations from each layer of the model. The best performance is observed at the 16th layer, reaching 70.8\%, whereas both shallower and deeper layers yield lower accuracy. This indicates that representations from intermediate layers are particularly effective for the anomaly detection. This characteristics can be explained by the role of each layer in the model. Early layers tend to encode low-level, local features, while final layers are often specialized toward the model’s pretraining objective. Intermediate layers, by contrast, capture general and abstract representations that are more transferable to the anomaly detection task. Similar findings have been reported in NLP~\cite{skean2025layer} and CV~\cite{damm2025anomalydino}. Our results confirm that this trend also holds in the time series anomaly detection.

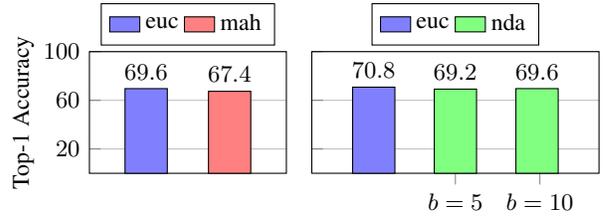
\begin{figure}[!t]
  % \centering
  \begin{subfigure}[b]{0.45\linewidth}
    % \centering
    \begin{tikzpicture}
        \begin{axis}[
            % ybar,
            ybar=-0.5cm,
            bar width=16pt,
            width=4.2cm,
            height=3.2cm,
            enlarge x limits=0.75,
            symbolic x coords={euc, mah},
            xtick={euc,mah},
            xticklabel style={text=white},
            xtick style={draw=none},
            ymin=0, ymax=100,
            ylabel={Top-1 Accuracy},  % ylabel={\hspace{1.8em}Top-1 Accuracy},
            ytick={20,60,100},
            legend style={at={(0.5,1.05)}, anchor=south, legend columns=2, font=\small},
            legend image code/.code={
                \draw[draw=black]
                (0cm,-0.1cm) rectangle (0.4cm,0.1cm);
            },
            ylabel style={yshift=-1em},
            grid=major,
            major x grid style={draw=none},
            ytick pos=left,
            tick label style={font=\small},
            label style={font=\small},
            nodes near coords,
            nodes near coords style={font=\small}
        ]
        \addplot[draw=black, fill=blue!50] coordinates {(euc, 69.6)}; \addlegendentry{euc}
        \addplot[draw=black, fill=red!50] coordinates {(mah, 67.4)}; \addlegendentry{mah}
        \end{axis}
    \end{tikzpicture}
  \end{subfigure}
  \hfill
  \begin{subfigure}[b]{0.7\linewidth}
    % \centering
    \begin{tikzpicture}
        \begin{axis}[
            % ybar,
            ybar=-0.6cm,
            bar width=16pt,
            width=5.4cm,
            height=3.2cm,
            enlarge x limits=0.35,
            symbolic x coords={euc, $b=5$, $b=10$},
            xtick={$b=5$, $b=10$},
            ylabel={\phantom{Top-1 Accuracy}},
            ymin=0, ymax=100,
            ytick={20,60,100},
            legend style={at={(0.5,1.05)}, anchor=south, legend columns=2, font=\small},
            legend image code/.code={
                \draw[draw=black]
                (0cm,-0.1cm) rectangle (0.4cm,0.1cm);
            },
            ylabel style={yshift=-5.0em},
            grid=major,
            major x grid style={draw=none},
            xtick pos=bottom,
            ytick pos=left,
            yticklabels={},
            tick label style={font=\small},
            label style={font=\small},
            nodes near coords,
            nodes near coords style={font=\small}
        ]
        \addplot[draw=black, fill=blue!50] coordinates {(euc, 70.8)}; \addlegendentry{euc}
        \addplot[draw=black, fill=green!50] coordinates {($b=5$, 69.2)}; \addlegendentry{nda}
        \addplot[draw=black, fill=green!50] coordinates {($b=10$, 69.6)};
        \end{axis}
    \end{tikzpicture}
  \end{subfigure}
  \caption{Top-1 accuracy using Euclidean, Mahalanobis, and nearest density-based distances.}
  \label{fig4}
\end{figure}

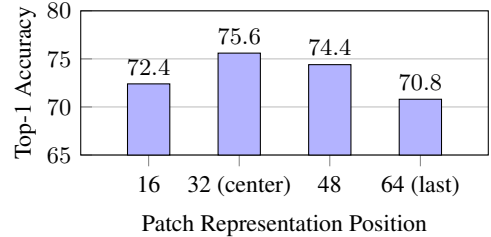
\begin{figure}
\centering
\begin{tikzpicture}
\begin{axis}[
    ybar,
    bar width=16pt,
    width=7cm,
    height=3.5cm,
    ymin=65, ymax=80,
    ylabel={Top-1 Accuracy},
    xlabel={Patch Representation Position},
    symbolic x coords={16, 32 (center), 48, 64 (last)},
    xtick=data,
    ytick={65,70,75,80},
    ylabel style={yshift=-1.4em},
    xlabel style={yshift=-0.4em},
    nodes near coords,
    nodes near coords style={font=\small},
    tick label style={font=\small},
    label style={font=\small},
    grid=major,
    major x grid style={draw=none},
    xtick pos=bottom,
    ytick pos=left,
    enlarge x limits=0.25,
]
\addplot[draw=black, fill=blue!30] coordinates {(16, 72.4) (32 (center), 75.6) (48, 74.4) (64 (last), 70.8)};
\end{axis}
\end{tikzpicture}
\caption{Top-1 accuracy across different patch positions, each corresponding to a reference time step.}
\label{fig5}
\end{figure}

\subsection{Distance Function Comparison}

We adopt a distance-based scoring mechanism for anomaly detection and evaluate its performance under different distance functions. The left panel of Figure~\ref{fig4} compares Top-1 accuracy using Euclidean and Mahalanobis distances. Although Mahalanobis accounts for the covariance structure, it requires matrix inversion, which can fail with limited training data. This issue affected 20 datasets in the UCR Anomaly Archive, so results are reported on the remaining 230. The right panel compares Euclidean distance with a density-based variant using nearest's $b$ neighbors. We report results for several values of $b$ to assess its impact. Overall, Euclidean distance delivers consistently strong accuracy, confirming its effectiveness as a default choice for anomaly scoring.

\subsection{Impact of Temporal Alignment within Subsequence}

In constructing input subsequences for anomaly detection, the position of the reference time step within the window can be varied. A common choice is the last-aligned subsequence, where the current time step $t$ corresponds to the final element of the window. Alternatively, a center-aligned subsequence places $t$ at the temporal center, allowing the window to include both past and future observations. This design choice affects the nature of the intermediate representations extracted from the model. Here, the reference time step determines which patch-level representation is used to compute the anomaly score. For example, when using a last-aligned subsequence, the final patch representation is selected; for a center-aligned subsequence, the center patch is used instead. To investigate how this alignment affects anomaly detection performance, we conduct a position-wise evaluation using different patch positions, each corresponding to a specific reference time step. As shown in Figure~\ref{fig5}, selecting the center patch at position 32 yields the highest Top-1 accuracy of 75.6\%, followed by earlier positions such as position 48. In contrast, the final patch at position 64, which corresponds to the last-aligned setting, results in lower performance with a Top-1 accuracy of 70.8\%. These results suggest that placing the reference time step at the temporal center provides richer contextual information, leading to more effective anomaly detection.

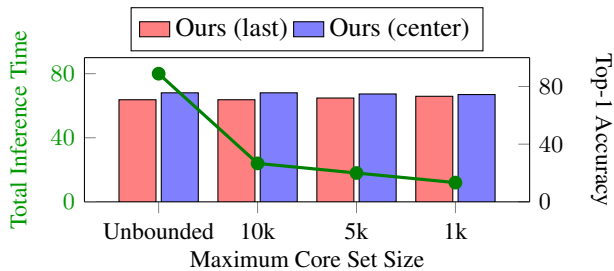
\begin{figure}
\begin{tikzpicture}
\begin{axis}[
    ybar,
    width=7.5cm,
    height=3.5cm,
    bar width=14pt,
    enlarge x limits=0.25,
    symbolic x coords={Unbounded,10k,5k,1k},
    xtick=data,
    ytick={0,40,80},
    axis y line*=right,
    ymin=0, ymax=100,
    ylabel={Top-1 Accuracy},
    xlabel={Maximum Core Set Size},
    ylabel style={
        color=black,
        at={(axis description cs:1.37,0.5)},
        rotate=180,
        font=\small
    },
    nodes near coords style={font=\small},
    tick label style={font=\small},
    label style={font=\small},
    yticklabel style={color=black},
    legend style={
        at={(0.5,1.35)},
        anchor=north,
        legend columns=2
    },
    legend image code/.code={
        \draw[draw=black](0cm,-0.1cm) rectangle (0.4cm,0.1cm);
    },
    xtick pos=bottom
]
\addplot[fill=red!50]  coordinates {(Unbounded,70.8) (10k,70.8) (5k,72.0) (1k,73.2)}; \addlegendentry{Ours (last)};
\addplot[fill=blue!50] coordinates {(Unbounded,75.6) (10k,75.6) (5k,74.8) (1k,74.4)}; \addlegendentry{Ours (center)};
\end{axis}
\begin{axis}[
    axis x line=none,
    axis y line*=left,
    width=7.5cm,
    height=3.5cm,
    enlarge x limits=0.25,
    ymin=0, ymax=90,
    ylabel={Total Inference Time},
    ylabel style={color=green!60!black,font=\small},
    ylabel style={yshift=-1em},
    yticklabel style={color=green!60!black},
    symbolic x coords={Unbounded,10k,5k,1k},
    xtick=data,
    ytick={0,40,80},
    nodes near coords style={font=\small},
    tick label style={font=\small},
    label style={font=\small}
]
  \addplot[green!50!black, very thick, mark=*] coordinates {(Unbounded,80) (10k,24) (5k,18) (1k,12)};
\end{axis}
\end{tikzpicture}
\caption{Top-1 accuracy and total inference time in hours across datasets under varying maximum core set sizes.}
\label{fig6}
\end{figure}

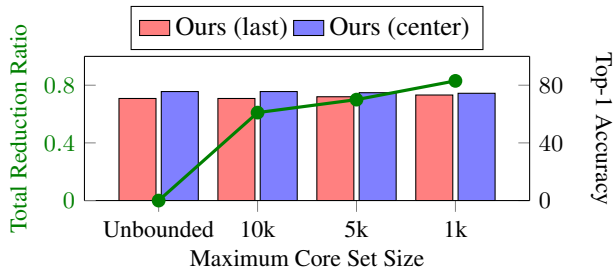
\begin{figure}
\begin{tikzpicture}
\begin{axis}[
    ybar,
    width=7.5cm,
    height=3.5cm,
    bar width=14pt,
    enlarge x limits=0.25,
    symbolic x coords={Unbounded,10k,5k,1k},
    xtick=data,
    axis y line*=right,
    ymin=0, ymax=100,
    ytick={0,40,80},
    ylabel={Top-1 Accuracy},
    xlabel={Maximum Core Set Size},
    ylabel style={
        color=black,
        at={(axis description cs:1.37,0.5)},
        rotate=180,
        font=\small
    },
    yticklabel style={color=black},
    legend style={
        at={(0.5,1.35)},
        anchor=north,
        legend columns=2
    },
    legend image code/.code={
        \draw[draw=black](0cm,-0.1cm) rectangle (0.4cm,0.1cm);
    },
    xtick pos=bottom,
    nodes near coords style={font=\small},
    tick label style={font=\small},
    label style={font=\small}
]
\addplot[fill=red!50]  coordinates {(Unbounded,70.8) (10k,70.8) (5k,72.0) (1k,73.2)}; \addlegendentry{Ours (last)};
\addplot[fill=blue!50] coordinates {(Unbounded,75.6) (10k,75.6) (5k,74.8) (1k,74.4)}; \addlegendentry{Ours (center)};
\end{axis}
\begin{axis}[
    axis x line=none,
    axis y line*=left,
    width=7.5cm,
    height=3.5cm,
    enlarge x limits=0.25,
    ymin=0, ymax=1,
    ylabel={Total Reduction Ratio},
    ylabel style={color=green!50!black,font=\small},
    ylabel style={yshift=-1em},
    yticklabel style={color=green!50!black},
    symbolic x coords={Unbounded,10K,5K,1K},
    xtick=data,
    ytick={0,0.4,0.8},
    nodes near coords style={font=\small},
    tick label style={font=\small},
    label style={font=\small}
]
  \addplot[green!50!black, very thick, mark=*] coordinates {(Unbounded,0) (10K,0.61) (5K,0.7) (1K,0.83)};
\end{axis}
\end{tikzpicture}
\caption{Top-1 accuracy and total reduction ratio across datasets under varying maximum core set sizes.}
\label{fig7}
\end{figure}

\subsection{Core Set Efficiency Evaluation}

As described in Section 3.3, we apply the greedy k-Center method to reduce the size of the memory bank before inference. We evaluate how this compression affects anomaly detection performance and computational efficiency by varying the maximum core set size.
As shown in Figure~\ref{fig6}, reducing the memory size from unbounded to 10k, 5k, and 1k entries leads to only a marginal decrease in Top-1 accuracy for both alignment settings. For instance, the center-aligned model maintains over 75\% accuracy even when compressed to 1k entries. Meanwhile, inference time decreases substantially as the memory bank becomes smaller, demonstrating the practical efficiency gained through coreset selection.
In Figure~\ref{fig7}, we further observe that the total reduction ratio increases steadily with smaller core set sizes, reaching over 80\% at the 1K limit. Despite this significant reduction in memory size, the detection accuracy remains stable. These findings highlight that the memory compression enables lightweight and scalable deployment while maintaining robust anomaly detection performance.

\begin{figure}
\centering
\begin{tikzpicture}
\begin{axis}[
    ybar,
    bar width=16pt,
    width=7cm,
    height=3.5cm,
    ymin=65, ymax=82,
    ylabel={Top-1 Accuracy},
    ytick={70,76},
    symbolic x coords={Ours (last), Ours (center)},
    xtick=data,
    legend style={at={(0.5,1.35)}, anchor=north, legend columns=2},
    legend image code/.code={
        \draw[draw=black](0cm,-0.1cm) rectangle (0.4cm,0.1cm);
    },
    nodes near coords,
    nodes near coords align={vertical},
    enlarge x limits=0.7,
    ylabel style={yshift=-1.2em},
    grid=major,
    xtick pos=bottom,
    ytick pos=left,
    major x grid style={draw=none},
    nodes near coords style={font=\small},
    tick label style={font=\small},
    label style={font=\small}
]
\addplot[
    draw=black, fill=red!50,
    line width=1pt,
] coordinates {(Ours (last), 70.8) (Ours (center), 75.6)};
\addplot[
    draw=black, pattern=north east lines, pattern color=blue, line width=1pt,
] coordinates {(Ours (last), 72.8) (Ours (center), 77.6)};
\legend{w/o TTAMB, w TTAMB}
\end{axis}
\end{tikzpicture}
\caption{Top‑1 accuracy with and without the test‑time adaptive memory bank method (TTAMB).}
\label{fig8}
\end{figure}
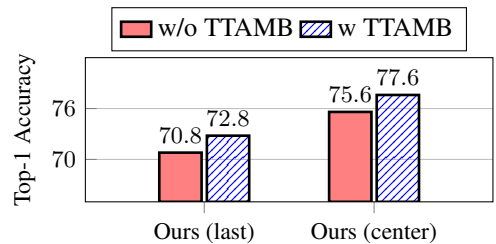

\subsection{Test-Time Adaptive Memory Bank Evaluation}

We evaluate the effectiveness of the test-time adaptive memory bank (TTAMB) introduced in Section 3.4. This mechanism augments the memory bank at inference time by incorporating non-redundant intermediate representations from incoming data, enabling adaptation to concept drift without updating model parameters.
The evaluation is conducted using the same benchmark as in previous sections, the UCR Anomaly Archive. This benchmark is particularly suitable for assessing TTAMB, as several datasets in the archive exhibit concept drift in their test sets.
Figure~\ref{fig8} presents the Top-1 accuracy with and without TTAMB for both last- and center-aligned configurations. In both settings, adaptation leads to consistent performance gains. Specifically, the center-aligned variant improves from 75.6\% to 77.6\%, and the last-aligned variant improves from 70.8\% to 72.8\%. These results confirm that TTAMB enhances robustness under concept drift.

\section{Conclusion}

We presented TimeRep, a simple yet effective anomaly detection method that leverages intermediate layer representations from a time series foundation model (TSFM). Unlike existing TSFM-based methods that perform anomaly detection through final layer representations and task-specific heads, TimeRep directly leverage intermediate layer computes anomaly scores based on the semantic distance of internal features, enabling robust performance without dataset-specific fine-tuning.
Through extensive experiments on the UCR Anomaly Archive, we demonstrated that TimeRep outperforms a wide range of baselines, including traditional, deep learning, and foundation model-based methods. We further showed that incorporating coreset-based memory compression and test-time adaptive updates improves both efficiency and robustness, particularly under concept drift.
In future work, we plan to extend TimeRep to multivariate time series settings and explore how different foundation model architectures affect the quality of intermediate representations for anomaly detection.

\bibliography{aaai2026}

\newpage

\appendix

\section{Implementation Details}

\subsection{Experimental Environment}

% 실험은 GPU 실험은 NVIDIA A100 80GB 1대
% Python 3.10.16
% PyTorch 2.4.1+cu121

We conducted all our experiments on a single NVIDIA A100 80GB GPU. We used Python 3.10.16 and PyTorch 2.4.1+cu121. Also, we used faiss-cpu 1.10.0 for similarity search.

\subsection{Dataset and Preprocessing}

% UCR Anomaly Archive
% 추가적인 전처리를 적용하지 않고, 공개된 벤치마크 데이터셋 사용함.
% 모든 실험에서는 UCR Anomaly Archive 벤치마크에서 제공되는 총 250개의 데이터셋을 모두 사용함.

We used the UCR Anomaly Archive with no additional preprocessing. All our experiments were conducted directly on the publicly available 250 datasets provided in the UCR Anomaly Archive.

\subsection{Experimental Settings}

% 기본적으로 모든 실험은 윈도우의 보폭을 1로 설정을 따름. 이 윈도우의 보폭은 학습과 추론에 관계 없이 모두 동일하게 1로 설정됨. 윈도우 길이는 각 방법에서 사용된 설정을 따름. 그렇기 때문에 어떤 방법은 250개 데이터셋에 대해 개별 데이터셋별로 윈도우의 길이가 다른 반면에 다른 방법은 250개 데이터셋에 대해 윈도우의 길이가 모두 동일할 수 있음. 

Table~\ref{app:tab1} lists the source of the results for the methods compared in Table~\ref{tab1} of the main paper. Table~\ref{app:tab2} provides the sources for the methods shown in Figure~\ref{fig2}. All methods are evaluated using a sliding window with a stride of 1, consistently applied during both training and inference to ensure fair comparisons across time series anomaly detection scenarios. Although the stride is fixed across all experiments, the window length varies depending on each method's original configuration. Consequently, some methods use a fixed window length across all 250 datasets, whereas others employ dataset-specific window lengths.

\begin{table}[!ht]
\centering
\renewcommand{\arraystretch}{1.2}
%\resizebox{.95\columnwidth}{!}{
\begin{tabular}{c|c}
    \hline
    \textbf{Method} & \textbf{Source} \\
    \hline
    IForest & Reported \cite{audibert2021univariate} \\
    SCRIMP & Reported \cite{lu2022matrix} \\
    NORMA & Reported \cite{lu2022matrix} \\
    DAMP & Reported \cite{lu2022matrix} \\
    LSTM-VAE & Reported \cite{lu2022matrix} \\
    AE & Reported \cite{audibert2021univariate} \\
    USAD & Reported \cite{audibert2021univariate} \\
    COCA & Reproduced (from this paper) \\
    TimeVQVAE-AD & Reproduced (from this paper) \\
    MOMENT & Reproduced (from this paper) \\
    Timer & Reproduced (from this paper) \\
    \hline
\end{tabular}
\caption{Sources of the results in Table~\ref{tab1} of the main paper. Methods labeled as \textit{Reported} use performance values from the original publications, while those labeled as \textit{Reproduced} are implemented and evaluated in this study.}
\label{app:tab1}
\end{table}

\begin{table}[!ht]
\centering
\renewcommand{\arraystretch}{1.2}
%\resizebox{.95\columnwidth}{!}{
\begin{tabular}{c|c}
    \hline
    \textbf{Method} & \textbf{Source} \\
    \hline
    Timer & Reported \cite{liu2024timer} \\
    DADA & Reported \cite{shentu2025towards} \\
    \hline
\end{tabular}
\caption{Sources of the results in Table \ref{fig2} of the main paper. Methods marked as \textit{Reported} use performance values from the original publications.}
\label{app:tab2}
\end{table}

\subsubsection{TimeRep (Ours)}

% TimeRep 실험 설정
% 모든 실험에서는 윈도우 길이 512로 윈도우 보폭을 1로 설정함.
% 윈도우 길이 512는 MOMENT 논문에서 MOMENT-Large 모델을 사전 학습시킬 때 사용한 값이며, 본 연구에서도 그대로 사용함. 사전 학습에 학습된 특징 표현을 최대한 이용하는 것을 목표로.

For all our experiments, TimeRep use a fixed window length of 512 and a stride of 1. The choice of window length follows the configuration used to pretrain the MOMENT-Large model in the original MOMENT paper. We adopt the same setting to fully leverage the pretrained representations learned in that work.

\subsubsection{COCA}

% COCA는 재현 가능한 코드와 실험 설정을 공개하고 있음. 우리는 이 공개된 코드와 실험 설정을 사용함. 실험 설정에서 윈도우 스트라이드만 1로 설정하여 UCR Anomaly Archive의 250개에 대해 개별 데이터셋에 대해 학습과 평가를 수행함. 재현 대상은 데이터 증강을 적용하지 않는 방법을 재현함. 

COCA provides publicly available code and experimental settings, which we adopt in our experiments. We follow their original setup with one modification: the window stride is set to 1 during both training and evaluation. All experiments are conducted individually on each of the 250 datasets from the UCR Anomaly Archive. We reproduce the version of COCA that does not apply data augmentation.

\subsubsection{TimeVQVAE-AD}

% TimeVQVAE-AD는 재현 가능한 코드와 실험 설정을 공개하고 있음. 우리는 이 공개된 코드와 실험 설정을 변경하지 않고 그대로 사용함. UCR Anomaly Archive의 250개에 대해 개별 데이터셋에 대해 학습과 평가를 수행함.

TimeVQVAE-AD provides publicly available code and experimental settings, which we use with no modifications. We train and evaluate the model separately on each of the 250 datasets from the UCR Anomaly Archive.

\subsubsection{Timer}

% Timer는 구현 코드와 실험 설정을 공개하고 있음. 윈도우 스트라이드를 1로 수정함. Timer 논문에서는 \alpha-quantile 평가 결과를 보고하고 있음. 우리는 논문에 보고된 \alpha-quantile 평가 결과를 사용함. 또한, Top-1 Accuracy 평가 지표를 구현하여 평가함.

Timer provides publicly available implementation and experimental configurations, which we adopt with a minor modification: the window stride is set to 1 to align with our evaluation protocol. The original Timer paper reports results using the $\alpha$-quantile metric, which we use as reported. Additionally, we implement and evaluate the model using the Top-1 accuracy metric.

\subsubsection{MOMENT}

% MOMENT는 구현 코드와 실험 설정을 공개하고 있음. 우리는 이 공개된 코드와 실험 설정을 기반으로 실험을 수행함. MOMENT 논문에서는 PA 기반 F1-score와 VUS-ROC를 평가 지표로 사용하며, 윈도우 보폭을 non-overlapping 방식으로 사용함. 우리는 본 연구에서 사용한 실험에 맞추기 위해, Top-1 aucrracy 평가 지표를 추가하고, 윈도우 보폭을 1로 하여 실험을 수행함.

MOMENT provides publicly available code and experimental settings, which we use as the foundation for our experiments. The original paper evaluates performance using PA-based F1-score and VUS-ROC metrics with non-overlapping windows. To align with the evaluation protocol adopted in this study, we modify the window stride to 1 and additionally evaluate the model using the Top-1 accuracy metric.

\end{document}